# Refractor Importance Sampling

**Haohai Yu** and **Robert A. van Engelen**
Department of Computer Science
Florida State University
Tallahassee, FL 32306-4530 USA
{hyu,engelen}@cs.fsu.edu

## Abstract

In this paper we introduce Refractor Importance Sampling (RIS), an improvement to reduce error variance in Bayesian network importance sampling propagation under evidential reasoning. We prove the existence of a collection of importance functions that are close to the optimal importance function under evidential reasoning. Based on this theoretic result we derive the RIS algorithm. RIS approaches the optimal importance function by applying localized arc changes to minimize the divergence between the evidence-adjusted importance function and the optimal importance function. The validity and performance of RIS is empirically tested with a large set of synthetic Bayesian networks and two real-world networks.

## 1 Introduction

The Bayesian Network (BN) [Pearl, 1988] formalism is one of the dominant representations for modeling uncertainty in intelligent systems [Neapolitan, 1990, Russell and Norvig, 1995]. A BN is a probabilistic graphical model of a joint probability distribution over a set of statistical variables. Bayesian inference on a BN answers probabilistic queries about the variables and their influence relationships. The posterior probability distribution is computed using belief updating methods [Pearl, 1988, Guo and Hsu, 2002]. Exact inference is NP-hard [Cooper, 1990]. Thus, exact methods only admit relatively small networks or simple network configurations in the worst case. Approximations are also NP-hard [Dagum and Luby, 1993]. However, approximate inference methods have anytime [Garvey and Lesser, 1994] and/or anywhere [Santos et al., 1995] properties that make these methods more attractive compared to exact methods.

Stochastic simulation algorithms, also called stochastic sampling or Monte Carlo (MC) algorithms, form one of the most prominent subclasses of approximate inference algorithms of which Logic Sampling [Henrion, 1988] was the first and simplest sampling algorithm. Likelihood weighting [Fung and Chang, 1989] was designed to overcome the poor performance of logic sampling under evidential reasoning with unlikely evidence. Markov Chain Monte Carlo (MCMC) forms another important group of stochastic sampling algorithms. Examples in this group are Gibbs sampling, Metropolis sampling and hybrid-MC sampling [Geman and Geman, 1984, Gilks et al., 1996, MacKay, 1998, Pearl, 1987, Chavez and Cooper, 1990]. Stratified sampling [Bouckaert, 1994], hypercube sampling [Cheng and Druzdzel, 2000c], and quasi-MC methods [Cheng and Druzdzel, 2000b] generate random samples from uniform distributions using various methods to improve sampling results. The importance sampling methods [Rubinstein, 1981] are widely used in Bayesian inference. Self Importance Sampling (SIS) [Shachter and Peot, 1990] and Adaptive Importance Sampling (AIS-BN) [Cheng and Druzdzel, 2000a] are among the most effective algorithms.

In this paper we prove that the importance functions of an evidence-updated BN can only approach the optimal importance function when the BN graph structure is modified according to the observed evidence. This implies the existence of a collection of importance functions with minimum divergence to the optimal importance function under evidential reasoning. Based on this result we derive our Refractor Importance Sampling (RIS) class of algorithms. In contrast to AIS-BN and SIS methods, RIS removes the lower bound that prevents the updated importance function to approach the optimal importance function. This is achieved by a graphical structure "refractor", consisting of a localized network structure change that minimizes the divergence between the evidence-adjusted importance function and the optimal importance function.

The remainder of this paper is organized as follows. Section 2 proves the existence of a lower bound on the divergence to the optimal importance function under evidential reasoning with a BN. The lower bound is used to derive the class of RIS algorithms introduced in Section 3. Section 4 empirically verifies the properties of the RIS algorithms on a large set of synthetic networks and two real-world networks, and compares the results to other importance sampling algorithms. Finally, Section 5 summarizes our conclusions and describes our future work.

## 2 Importance Function Divergence

In this section we first give BN definitions and briefly review importance sampling. We then give a KL-divergence lower bound for importance sampling error variance. We prove the existence of a collection of importance functions that approach the optimal importance function by adjusting both the quantitative and qualitative components of a BN under dynamic updating with evidence.

### 2.1 Definitions

The following definitions and notations are used.

**Def. 1** *A Bayesian network $BN = (G, \Pr)$ is a DAG $G = (\mathbf{V}, \mathbf{A})$ with vertices $\mathbf{V}$ and arcs $\mathbf{A}$, $\mathbf{A} \subseteq \mathbf{V} \times \mathbf{V}$. $\Pr$ is the joint probability distribution over the discrete random variables (vertices) $\mathbf{V}$ defined by $\Pr(\mathbf{V}) = \prod_{V \in \mathbf{V}} \Pr(V \mid \pi(V))$. The set of parents of a vertex $V$ is $\pi(V)$. The conditional probability tables (CPT) of the BN assign values to $\Pr(V \mid \pi(V))$ for all $V \in \mathbf{V}$.*

The graph $G$ induces the *d-separation criterion* [Pearl, 1988], denoted by $\langle \mathbf{X}, \mathbf{Y} \mid \mathbf{Z} \rangle$, which implies that $\mathbf{X}$ and $\mathbf{Y}$ are conditionally independent in $\Pr$ given $\mathbf{Z}$, with $\mathbf{X}, \mathbf{Y}, \mathbf{Z} \subseteq \mathbf{V}$.

**Def. 2** *Let $BN = (G, \Pr)$ be a Bayesian network.*

- *The combined parent set of $\mathbf{X} \subseteq \mathbf{V}$ is defined by $\boldsymbol{\pi}(\mathbf{X}) = \bigcup_{X \in \mathbf{X}} \pi(X) \setminus \mathbf{X}$.*

- *Let $An(\cdot)$ denote the transitive closure of $\pi(\cdot)$, i.e. the ancestor set of a vertex. The combined ancestor set of $\mathbf{X} \subseteq \mathbf{V}$ is defined by $\mathbf{An}(\mathbf{X}) = \bigcup_{X \in \mathbf{X}} An(X) \setminus \mathbf{X}$.*

- *Let $\delta : \mathbf{V} \to \mathbb{N}$ denote a topological order of the vertices such that $Y \in An(X) \to \delta(Y) < \delta(X)$. The ahead set of a vertex $X \in \mathbf{V}$ given $\delta$ is defined by $Ah(X) = \{Y \in \mathbf{V} \mid \delta(Y) < \delta(X)\}$.*

### 2.2 Importance Sampling

Importance sampling is an MC method to improve the convergence speed and reduce the error variance with probability density functions. Let $g(\mathbf{X})$ be a function of $m$ variables $\mathbf{X} = \{X_1, \ldots, X_m\}$ over domain $\Omega \subseteq \mathbb{R}^m$, such that computing $g(\mathbf{X})$ for any $\mathbf{X}$ is feasible. Consider the problem of approximating $I = \int_\Omega g(\mathbf{X}) d\mathbf{X}$ using a sampling technique. Importance sampling approaches this problem by rewriting $I = \int_\Omega \frac{g(\mathbf{X})}{f(\mathbf{X})} f(\mathbf{X}) d\mathbf{X}$, where $f(\mathbf{X})$ is a probability density function over $\Omega$, often referred to as the importance function. In order to achieve minimum error variance equal to $\sigma^2_{f(\mathbf{X})} = (\int_\Omega |g(\mathbf{X})| d\mathbf{X})^2 - I^2$, the importance function should be $f(\mathbf{X}) = |g(\mathbf{X})|(\int_\Omega |g(\mathbf{X})| d\mathbf{X})^{-1}$, see [Rubinstein, 1981]. Note that when $g(\mathbf{X}) > 0$ the optimal probability density function is $f(\mathbf{X}) = g(\mathbf{X})I^{-1}$ and $\sigma^2_{f(\mathbf{X})} = 0$. It is obvious that in most of cases it is impossible to obtain the optimal importance function.

The SIS [Shachter and Peot, 1990] and AIS-BN [Cheng and Druzdzel, 2000a] sampling algorithms are effective methods for approximate Bayesian inference. These methods attempt to approach the optimal importance function through learning by dynamically adjusting the importance function during sampling with evidence. To this end, AIS-BN heuristically changes the CPT values of a BN, a technique that has been shown to significantly improve the convergence rate of the approximation to the exact solution.

We use the following definitions for sake of exposition.

**Def. 3** *Let $BN = (G, \Pr)$ be a Bayesian network with $G = (\mathbf{V}, \mathbf{A})$ and evidence $\mathbf{e}$ for variables $\mathbf{E} \subseteq \mathbf{V}$. A posterior $BN_\mathbf{e}$ of the BN is some (new) network defined as $BN_\mathbf{e} = (G_\mathbf{e}, \Pr_\mathbf{e})$ with graph $G_\mathbf{e}$ over variables $\mathbf{V} \setminus \mathbf{E}$, such that $BN_\mathbf{e}$ exactly models the posterior joint probability distribution $\Pr_\mathbf{e} = \Pr(\cdot \mid \mathbf{e})$.*

A typical example of a *posterior $BN_\mathbf{e}$* is a BN combined with an updated posterior state as defined by exact inference algorithms, e.g. using evidence absorption [van der Gaag, 1996]. Approximations of $BN_\mathbf{e}$ are used by importance sampling algorithms. These approximations consist of the original BN with all evidence vertices ignored from further consideration.

**Def. 4** *Let $BN = (G, \Pr)$ be a Bayesian network with $G = (\mathbf{V}, \mathbf{A})$ and evidence $\mathbf{e}$ for variables $\mathbf{E} \subseteq \mathbf{V}$. The evidence-simplified $ESBN_\mathbf{e}$ of BN is defined by $ESBN_\mathbf{e} = (G'_\mathbf{e}, \Pr'_\mathbf{e})$, where $G'_\mathbf{e} = (\mathbf{V}'_\mathbf{e}, \mathbf{A}'_\mathbf{e})$, $\mathbf{V}'_\mathbf{e} = \mathbf{V} \setminus \mathbf{E}$, and $\mathbf{A}'_\mathbf{e} = \{(X, Y) \mid (X, Y) \in \mathbf{A} \bigwedge X, Y \notin \mathbf{E}\}$.*

The joint probability distribution $\Pr'_\mathbf{e}$ of an evidence-simplified BN approximates $\Pr_\mathbf{e}$. For example, SIS and AIS-BN adjust the CPTs of the original BN.

## 2.3 KL-Divergence Bounds

We give a lower bound on the KL-divergence [Kullback, 1959] of the evidence-simplified $\Pr'_{\mathbf{e}}$ from the exact $\Pr_{\mathbf{e}}$. The lower bound is valid for all variations of $\Pr'_{\mathbf{e}}$, including those generated by importance sampling algorithms that adjust the CPT.

**Theorem 1** *Let $ESBN_{\mathbf{e}} = (G'_{\mathbf{e}}, \Pr'_{\mathbf{e}})$ be an evidence-simplified BN given evidence $\mathbf{e}$ for $\mathbf{E} \subseteq \mathbf{V}$. If $\Pr'_{\mathbf{e}}(V \mid \pi_{\mathbf{e}}(V)) = \Pr(V \mid \pi(V), \mathbf{e})$ for all $V \in \mathbf{V}$ then the KL-divergence between $\Pr_{\mathbf{e}}$ and $\Pr'_{\mathbf{e}}$ is minimal and given by*

$$\sum_{X \in \mathbf{X}} \sum_{Cfg(X, \pi(X))} \Pr(x, \pi(x) \mid \mathbf{e}) \ln \Pr(x \mid \pi(x)) +$$
$$\sum_{X \in \mathbf{X}} \sum_{Cfg(X, \pi(X))} \Pr(x, \pi(x) \mid \mathbf{e}) \ln \frac{1}{\Pr'_{\mathbf{e}}(x \mid \pi_{\mathbf{e}}(x))} +$$
$$\sum_{Cfg(\boldsymbol{\pi}(\mathbf{E}))} \Pr(\boldsymbol{\pi}(\mathbf{e}) \mid \mathbf{e}) \ln \prod_{e \in \mathbf{e}} \Pr(e \mid \pi(e)) - \ln \Pr(\mathbf{e}) \quad (1)$$

*where $\mathbf{X} = \mathbf{V} \setminus \mathbf{E}$.*

**Proof.** See Appendix A. □

Theorem 1 bounds the error variance from below, which is empirically verified for SIS and AIS-BN in the results Section 4. The divergence Eq. (1) is zero when specific conditions are met as stated below.

**Corollary 1** *Let $ESBN_{\mathbf{e}} = (G'_{\mathbf{e}}, \Pr'_{\mathbf{e}})$ be an evidence-simplified BN given evidence $\mathbf{e}$ for $\mathbf{E} \subseteq \mathbf{V}$. If $\boldsymbol{\pi}(\mathbf{E}) \cap (\mathbf{V} \setminus \mathbf{E}) = \emptyset$, then $\Pr'_{\mathbf{e}} = \Pr_{\mathbf{e}}$.*

**Proof.** See Appendix B. □

Hence, the optimal importance function is obtained when all evidence vertices are clustered as roots in $G$.

We will now show how $\Pr'_{\mathbf{e}}$ can approach the optimal $\Pr_{\mathbf{e}}$ without restrictions. For sake of explanation, the following widely-held assumptions are reiterated:

**Assumption 1** *The topological order $\delta$ of a BN and its posterior version $\delta_{\mathbf{e}}$ of $BN_{\mathbf{e}}$ are consistent. That is, $\delta_{\mathbf{e}}(Y) < \delta_{\mathbf{e}}(X) \rightarrow \delta(Y) < \delta(X)$ for all $X, Y \in \mathbf{V} \setminus \mathbf{E}$.*

Assumption 1 is reasonable for the following facts:

1. According to chain rule, a BN can be built up in any topological order and all of them describe the same joint probability distribution.

2. Although there has never been a widely accepted definition of what causality is, it is widely accepted that the fact of observing evidence for random variables should not change the causality relationship between the variables.

**Theorem 2** *Let $BN_{\mathbf{e}}(G_{\mathbf{e}}, \Pr_{\mathbf{e}})$ be the posterior of a $BN = (G, \Pr)$ given evidence $\mathbf{e}$ for $\mathbf{E} \subseteq \mathbf{V}$. If $X \notin \mathbf{An}(\mathbf{E})$ for all $X \in \mathbf{V} \setminus \mathbf{E}$, then $\Pr_{\mathbf{e}}(X \mid Ah_{\mathbf{e}}(X)) = \Pr(X \mid \pi(X))$. The evidence vertices in $\pi(X)$ take configurations fixed by $\mathbf{e}$, that is $\Pr(X \mid \pi(X)) = \Pr(X \mid \pi(X) \setminus \mathbf{E}, e_1, \ldots, e_m)$ for all $e_i \in \pi(X) \cap \mathbf{E}$.*

**Proof.** See [Cheng and Druzdzel, 2000a]. □

Hence, to compute the posterior probability of a vertex that is not an ancestor of an evidence vertex, there is no need to change the parents of the vertex or its CPT. For vertices that are ancestors of evidence vertices, we use Bayes' formula and d-separation to explore the effects of evidence on those vertices. Without loss of generality, only one evidence vertex is considered. The result applies to an evidence vertex set by transitivity.

**Lemma 1** *Let $BN_{\mathbf{e}}(G_{\mathbf{e}}, \Pr_{\mathbf{e}})$ be the posterior of a $BN = (G, \Pr)$ given evidence $\mathbf{e} = \{e\}$ for $E \in \mathbf{V}$. Let $X \in An(E)$. Then, $\Pr_{\mathbf{e}}(X \mid Ah_{\mathbf{e}}(X)) = \frac{\Pr(e \mid X, Ah(X))}{\Pr(e \mid Ah(X))} \Pr(X \mid \pi(X))$.*

**Proof.** Because $Ah_{\mathbf{e}}(X) = Ah(X)$ by Assumption 1, we have $\Pr_{\mathbf{e}}(X \mid Ah_{\mathbf{e}}(X)) = \frac{\Pr_{\mathbf{e}}(X, Ah(X))}{\Pr_{\mathbf{e}}(Ah(X))} = \frac{\Pr(X, Ah(X) \mid e)}{\Pr(Ah(X) \mid e)} = \frac{\Pr(X, Ah(X), e)}{\Pr(Ah(X), e)} = \frac{\Pr(e \mid X, Ah(X)) \Pr(X, Ah(X))}{\Pr(e \mid Ah(X)) \Pr(Ah(X))} = \frac{\Pr(e \mid X, Ah(X))}{\Pr(e \mid Ah(X))} \Pr(X \mid \pi(X))$ by using Theorem 2. □

Theorem 2 and Lemma 1 show that if we have $\Pr(e \mid X, Ah(X))$ and $\Pr(e \mid Ah(X))$ for all $X \in An(E)$ we can derive $\Pr_{\mathbf{e}}(X \mid Ah_{\mathbf{e}}(X))$ to compute $\Pr_{\mathbf{e}}(\mathbf{V}) = \prod_{V \in An(E)} \Pr(V \mid \pi(V)) \prod_{V \notin An(E)} \Pr_{\mathbf{e}}(V \mid Ah_{\mathbf{e}}(V))$ for the optimal importance function. However, there are two problems to derive $\Pr_{\mathbf{e}}$. Firstly, $Ah(X)$ is too large to construct a *posterior* $\mathbf{BN}_{\mathbf{e}}$ for $\Pr_{\mathbf{e}}$ in practice. Secondly, instead of the exact $\Pr_{\mathbf{e}}(X \mid Ah_{\mathbf{e}}(X))$ we have an estimate by importance sampling.

The parent sets of $X \in An(E)$ can be minimized by exploiting d-separation. Let $\alpha_e(X) \subseteq X \cup Ah(X)$ denote the minimal vertex set that d-separates evidence $E$ and $X \cup Ah(X)$, thus $\langle E, X \cup Ah(X) \mid \alpha_e(X)\rangle$. We will refer to $\alpha_e(X)$ as the "shield" of $X$ given $E$. Let $\beta_e(X) \subseteq Ah(X)$ denote the minimal vertex set that d-separates evidence $E$ and $Ah(X)$, thus $\langle E, Ah(X) \mid \beta_e(X)\rangle$. We explore the relationship between $\alpha_e(X)$ and $\beta_e(X)$ below.

**Lemma 2** *Let $BN_{\mathbf{e}}(G_{\mathbf{e}}, \Pr_{\mathbf{e}})$ be the posterior of a $BN = (G, \Pr)$ given evidence $\mathbf{e} = \{e\}$ for $E \in \mathbf{V}$. Then, $\beta_e(X) \subseteq (\alpha_e(X) \setminus X) \cup \pi(X)$ for all $X \in An(E)$.*

**Proof.** See Appendix C. □

Therefore, we can approach the optimal importance function $\Pr_{\mathbf{e}}(X \mid Ah_{\mathbf{e}}(X))$ by estimation of $\Pr_{\mathbf{e}}(X \mid (\alpha_e(X) \setminus X) \cup \pi(X))$ from importance samples.

**Input**: Evidence $E \in \mathbf{V}$ and $X \in An(E)$
**Output**: The set $S = \alpha_e(X)$
**Data**: array $A$, queue $Q$
$A \leftarrow \text{topSort}_\delta(Ah(X))$;
$S \leftarrow \{X\}$;
**for** $i \leftarrow |A|$ **to** $1$ **do**
    $Q \leftarrow \emptyset$;
    push($Q$, $A[i]$);
    **while** $Q \neq \emptyset$ **do**
        $V \leftarrow \text{pop}(Q)$;
        **if** $V = E$ **then** $S \leftarrow S \cup \{A[i]\}$; **break**;
        **if** $V \notin S \wedge V \in An(E) \wedge X \notin An(V)$ **then**
            push(children($V$));
        **end**
    **end**
**end**

**Algorithm 1**: Computing the Shield $\alpha_e(X)$

**Input**: $BN = (G, \Pr)$, evidence $\mathbf{e}$ for $\mathbf{E} \subseteq \mathbf{V}$
**Output**: refractored $BN_\mathbf{e}$
**foreach** $E \in \mathbf{E}$ **do**
    **foreach** $X \in An(E)$ **do**
        expand $\pi_\mathbf{e}(X) = (\alpha_e(X) \setminus \{X\}) \cup \pi(X)$;
        update the CPT of $X$;
    **end**
**end**

**Algorithm 2**: Refractor Procedure

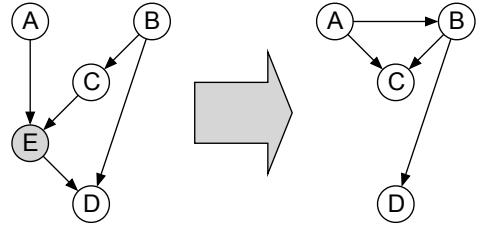

Figure 1: Refractor Example

## 3 Refractor Importance Sampling

The RIS algorithm modifies the BN structure according to the shield $\alpha_e(X)$ for vertices $X \in \mathbf{An(E)}$ by expanding the parent set of $X$ and adjusting its CPT accordingly. Visually in the graph, *RIS refracts arcs from the evidence vertices*, which inspired the choice of name for the method. The algorithms and general procedure of RIS are introduced in this section.

### 3.1 Computing the Shield

Alg. 1 computes $\alpha_e(X)$ in $O(|\mathbf{A}|)$ worst-case time, assuming $An(\cdot)$ is determined in unit time (e.g. using a lookup table). Function topSort$_\delta$ topologically sorts the set $Ah(X)$ by topological order $\delta$ over $\mathbf{V}$ of the BN. Note that the shield $\alpha_e(X)$ can be computed in advance for each $X \in \mathbf{V}$ given evidence nodes $\mathbf{E}$.

### 3.2 Refractor Procedure

Alg. 2 modifies the graphical structure of BN. The time complexity of this algorithm is $O(|\mathbf{V}||\mathbf{A}|)$ if $|\mathbf{E}| \ll |\mathbf{V}|$, otherwise it is $O(|\mathbf{V}|^2|\mathbf{A}|)$. The CPT of a vertex $X$ is updated by populating the expanded entries $\alpha_e(X) \setminus \{X\}$ using sampling data (described in Section 3.4).

Fig. 1 shows an example refractored BN using Alg. 2. E is the evidence node. Here, $\alpha_e(\mathsf{C}) = \{\mathsf{A}\}$ and $\alpha_e(\mathsf{B}) = \{\mathsf{A}\}$. Arcs $\mathsf{A} \to \mathsf{B}$ and $\mathsf{A} \to \mathsf{C}$ are added. Note that arc $\mathsf{A} \to \mathsf{B}$ adjusts for the fact that the influence relationship between A and B has changed through evidence E. Arc $\mathsf{E} \to \mathsf{D}$ is no longer required and can be removed as in [van der Gaag, 1996].

### 3.3 General RIS Procedure

RIS utilizes both the qualitative and quantitative properties of a BN to approach the optimal importance function. The general procedure of RIS is:

1. The structure of the BN is modified by Algorithms 1 and 2. The CPTs of (a subset of) ancestor vertices of evidence vertices are expanded.

2. Update the CPT values through some specific learning algorithm (see Section 3.4 for details).

3. Sample the BN with an importance sampling algorithm using the new importance function.

### 3.4 Variations of RIS

Step 1 modifies the BN structure significantly, especially when the ancestor sets of evidence vertices are large, e.g. when evidence vertices are leafs. This increases the complexity of the BN. However, the effect of evidence on other vertices is attenuated when the path length between the evidence and the vertices is increased [Henrion, 1989]. Therefore, instead of modifying all ancestors $\mathbf{An(E)}$ of evidence $\mathbf{E}$ in Step 1, it is generally sufficient to select a subset of ancestors such as the combined parent set $\boldsymbol{\pi}(\mathbf{E})$.

Steps 1 and 2 are independent, because any importance function learning algorithm can be applied in Step 2. Steps 2 and 3 can be combined by using the same importance sampling algorithm for learning and inference. In our experiments, we used SIS and AIS-BN for both learning and inference (steps 2 and 3), referred to as RISSIS and RISAIS, respectively. AIS-BN will be referred to by AIS.

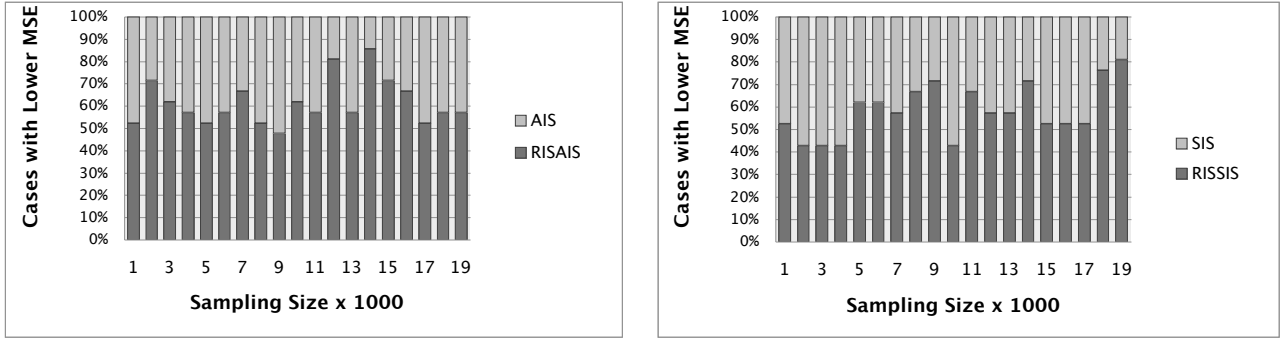

Figure 2: Synthetic BN Results: Ratio of Lowest MSE for RISAIS Versus AIS, and RISSIS Versus SIS.

## 4 Results

This section presents the experimental results of RISSIS and RISAIS compared to SIS and AIS for synthetic networks and two real-world networks.

### 4.1 Measurement

The *MSE* (mean squared error) metric was used to measure the error of the importance sampling results compared to the exact solution:

$$MSE = \sqrt{\frac{1}{\sum_{X_i \in \mathbf{X}} n_i} \sum_{X_i \in \mathbf{X}} \sum_{j=1}^{n_i} (\Pr'_\mathbf{e}(x_{ij}) - \Pr_\mathbf{e}(x_{ij}))^2} \ ,$$

where $\mathbf{X} = \mathbf{V} \setminus \mathbf{E}$. We also measured the KL-divergence of the approximate and exact posterior probability distributions:

$$KL\text{-}divergence = \sum_{Cfg(\mathbf{X})} \Pr_\mathbf{e}(\mathbf{x}) \ln \frac{\Pr_\mathbf{e}(\mathbf{x})}{\Pr'_\mathbf{e}(\mathbf{x})} \ .$$

Recall that Theorem 1 gives a lower bound for the KL-divergence of the posterior probability distributions of SIS and AIS, which is indicated in the results by the *PostKLD* lower bound from Eq. (1).

The number of samples is taken as a measure of running time instead of CPU time in our experimental implementation. Recall that the overhead of RIS is fixed at startup when the evidence set can be predetermined. Furthermore, the RIS overhead is limited to collecting the updated CPT values during sampling (and learning in the case of RISAIS).

The reported sampling frequencies for AIS (and RISAIS) are for calculating the posterior results. Because AIS separates the importance function learning stage from the sampling stage, the actual number of samples taken for AIS (total sampling for importance function and sampling the results) is twice that of SIS. Recommended parameters [Cheng and Druzdzel, 2000a] are used in AIS and RISAIS.

### 4.2 Test Cases

Because computing the *MSE* is expensive and *PostKLD* is exponential in the number of vertices, small-sized synthetic BNs with random variables with two or three states and $|\mathbf{V}| = 20$ vertices and $|\mathbf{A}| = 30$ arcs were evaluated in our experiments. The CPT for each variable is randomly generated with uniform distribution for the probability interval $[0.1, 0.9]$ with bias for the extreme probabilities in intervals $(0, 0.1)$ and $(0.9, 1)$. For the experiments we generated 100 different synthetic BNs with these characteristics.

We also verified RIS with two real-world BNs: *Alarm-37* [Beinlich et al., 1989] and *HeparII-70* [Onisko, 2003]. The probability distributions of these networks are more extreme compared to the synthetic BNs. For each of the two BNs, 20 sets of evidence variables are randomly chosen, each with 10 evidence variables. For the *Alarm-37* and *HeparII-70* we choose to limit the refractoring to the parents nodes of the evidence set $\boldsymbol{\pi}(\mathbf{E})$ instead of $\mathbf{An}(\mathbf{E})$, see Section 3.4.

### 4.3 Results for Synthetic Test Cases

We compared the MSE of four algorithms, AIS, RISAIS, SIS, and RISSIS. For this comparison a selection of 21 BNs from the generated synthetic test case suite was made. The other 79 test cases have $PostKLD \leq 0.1$, which means according to Theorem 1 that the RIS advantage is limited.

Fig. 2 shows the results for the 21 synthetic BNs, where the sample frequency is varied from 1,000 to 19,000 in increments of 1,000. The dark column in the figures represent the ratio of lowest MSE cases for RISAIS versus AIS and RISSIS versus SIS. A ratio of 50% or higher indicates that the RIS algorithm has lower error variance than the non-RIS algorithm. For RISAIS this is the case for all but one of the 19 measurements taken In total, the MSE is lowest for RISAIS in 61.4% on average over all samples. For RISSIS this is the case for all but four of the 19 measurements taken. In total, the MSE is lowest for RISSIS in 58.4% on average over all samples.

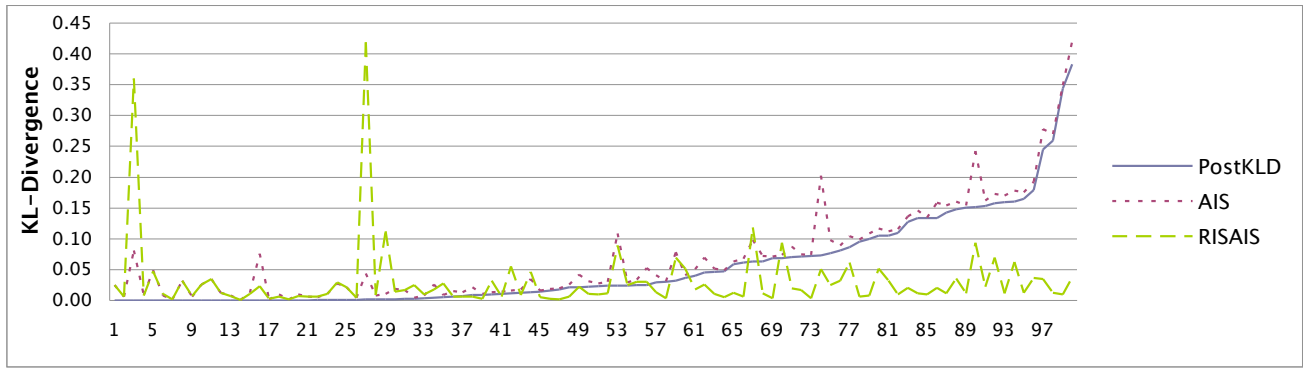

Figure 3: Synthetic BN Results: KL-Divergence of RISAIS and AIS with PostKLD Lower Bound

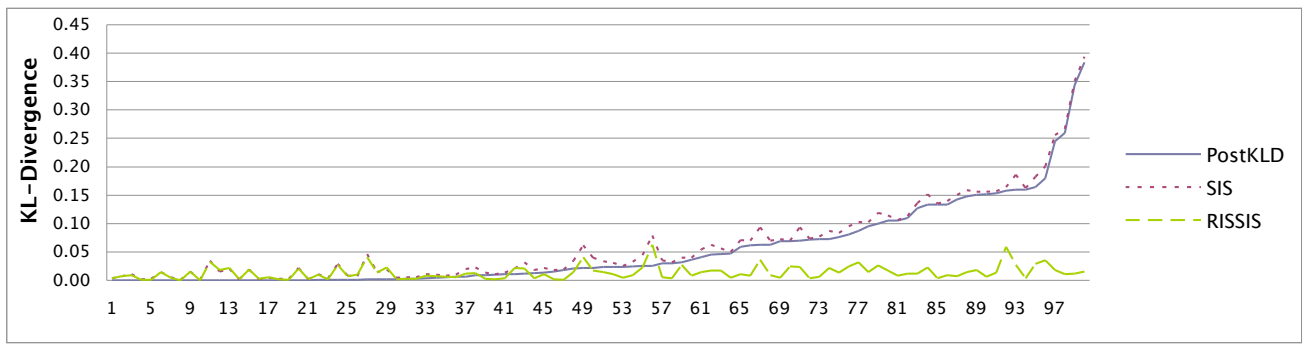

Figure 4: Synthetic BN Results: KL-Divergence of RISSIS and SIS with PostKLD Lower Bound

In fact, it is to be expected that the higher the *PostKLD* lower bound the better the RIS algorithms should perform. In order to determine the impact with increasing *PostKLD*, we selected all 100 synthetic BN test cases and measured the KL-divergence after 11,000 samples.

Fig. 3 shows the result for RISAIS, where the 100 BNs are ranked according to the *PostKLD*. Recall that the *PostKLD* is the lower bound on the KL-divergence of AIS. From the figure it can be concluded that AIS does not approach the exact solution for a significant number of test cases, whereas RISAIS is not limited by the bound due to the BN refractoring.

It should be noted that around points 1 and 26 in Fig. 3 the KL-divergence of RISAIS is worse compared to AIS. We believe the reason is that AIS heuristically changes the original CPT which has a negative impact on the RIS algorithm's ability to adjust the CPT to the optimal importance function.

Fig. 4 shows the result for RISSIS, where the 100 BNs are ranked according to the *PostKLD*. Interestingly, the RISSIS and SIS results are better on average than RISAIS and AIS. Note that the *PostKLD* lower bound is the same for AIS and SIS. However, in this study SIS appears to approach the *PostKLD* closer than AIS. Also here we can conclude that SIS does not approach the exact solution for a significant number of test cases, whereas RISSIS is not limited by the bound due to the BN refractoring.

### 4.4 Results for Alarm-37 and HeparII-70

Fig. 5 shows the results for *Alarm-37* and *HeparII-70*, where the sample frequency is varied from 1,000 to 19,000 in increments of 1,000. The dark column in the figures represent the ratio of lowest MSE cases for RISAIS versus AIS and RISSIS versus SIS. A ratio of 50% or higher indicates that the RIS algorithm has lower error variance than the non-RIS algorithm. For RISAIS this is the case for all but one of the 19 measurements taken. In total, the MSE is lowest for RISAIS in 56.7% on average over all samples. For RISSIS this is the case for all 19 measurements taken. In total, the MSE is lowest for RISSIS in 60.3% on average over all samples.

The combined results show that the RIS algorithms have reduced error variance for the synthetic networks and the two real-world networks. The *PostKLD* lower bound limits the ability of AIS and SIS to approach the optimal importance function. By contrast, the RIS approach successfully eliminates this limitation of AIS and SIS, thereby providing an improvement to reduce error variance in BN importance sampling propagation under evidential reasoning.

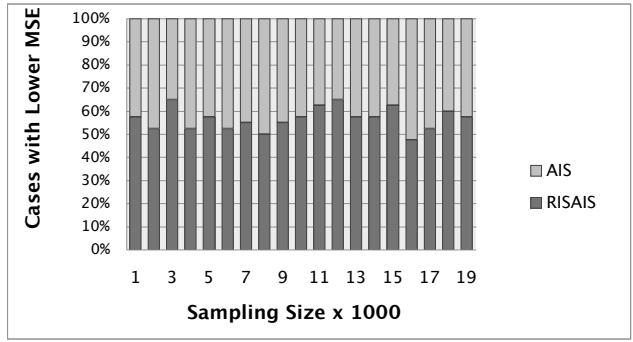 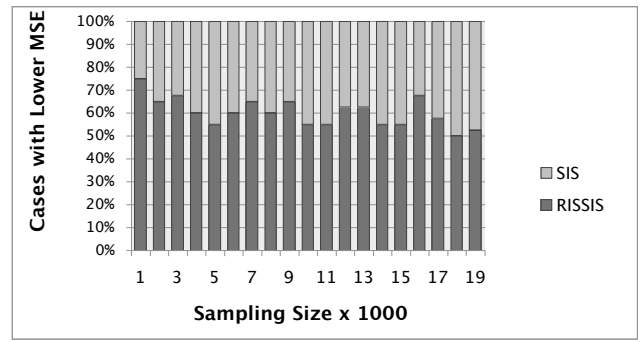

Figure 5: Ratio of Lowest MSE for RISAIS Versus AIS, and RISSIS Versus SIS (Alarm-37 and HeparII-70).

## 5 Conclusions

In order to approach the optimal importance function for importance sampling propagation under evidential reasoning with a Bayesian network, a modification of the network's structure is necessary to eliminate the lower bound on the error variance. To this end, the proposed RIS algorithms refractor the network and adjust the conditional probability tables to minimize the divergence to the optimal importance function. The validity and performance of the RIS approach was empirically tested with a set of synthetic networks and two real-world networks.

Additional improvements of RIS are possible to achieve a better accuracy/cost ratio. The goal is to find an effective subset of the full shield size of an ancestor vertex of an evidence vertex or select a limited subset of the ancestors of evidence vertices that are refractored. Also some of the additionally introduced arcs could be removed with the arc removal algorithm [van Engelen, 1997] when they present a weak influence. Such strategies would reduce the complexity of the refractored network while still ensuring higher accuracy over current importance sampling algorithms.

# Appendix

## A  Proof of Theorem 1

**Proof.** We use the KL-divergence (Cross Entropy) [Kullback, 1959] to measure the difference between the *posterior* $BN_\mathbf{e}$ and $ESBN_\mathbf{e}$. The KL-divergence = $\mathbf{E}_1[\ln \frac{\Pr_1(\mathbf{V})}{\Pr_2(\mathbf{V})}] = \sum_{Cfg(\mathbf{V})} \Pr_1(\mathbf{v}) \ln \frac{\Pr_1(\mathbf{v})}{\Pr_2(\mathbf{v})}$. Hence, the KL-divergence between *posterior* $BN_\mathbf{e}$ and $ESBN_\mathbf{e}$ is $\sum_{Cfg(\mathbf{X})} \Pr_\mathbf{e}(\mathbf{x}) \ln \frac{\Pr_\mathbf{e}(\mathbf{x})}{\Pr'_\mathbf{e}(\mathbf{x})}$ where $\mathbf{X} = \mathbf{V} \setminus \mathbf{E}$. This is further simplified as follows

$\sum_{Cfg(\mathbf{X})} \Pr_\mathbf{e}(\mathbf{x}) \ln \frac{\Pr_\mathbf{e}(\mathbf{x})}{\Pr'_\mathbf{e}(\mathbf{x})} =$
$\sum_{Cfg(\mathbf{X})} \Pr(\mathbf{x} \mid \mathbf{e}) \ln \frac{\Pr(\mathbf{x}\mid\mathbf{e})}{\Pr'_\mathbf{e}(\mathbf{x})} =$
$\sum_{Cfg(\mathbf{X})} \Pr(\mathbf{x} \mid \mathbf{e}) \ln \frac{\prod_{x_j\in\mathbf{x}}\Pr(x_j|\pi(x_j))\prod_{e_i\in\mathbf{e}}\Pr(e_i|\pi(e_i))}{\Pr(\mathbf{e})\prod_{x_j\in\mathbf{x}}\Pr'_\mathbf{e}(x_j|\pi_\mathbf{e}(x_j))}$
$= \sum_{Cfg(\mathbf{X})} \Pr(\mathbf{x} \mid \mathbf{e}) \ln \frac{\prod_{x_j\in\mathbf{x}}\Pr(x_j|\pi(x_j))\prod_{e_i\in\mathbf{e}}\Pr(e_i|\pi(e_i))}{\prod_{x_j\in\mathbf{x}}\Pr'_\mathbf{e}(x_j|\pi_\mathbf{e}(x_j))}$
$+ \ln \frac{1}{\Pr(\mathbf{e})} \sum_{Cfg(\mathbf{X})} \Pr(\mathbf{x} \mid \mathbf{e}) =$
$\sum_{Cfg(\mathbf{X})} \Pr(\mathbf{x} \mid \mathbf{e}) \ln \frac{\prod_{x_j\in\mathbf{x}}\Pr(x_j|\pi(x_j))\prod_{e_i\in\mathbf{e}}\Pr(e_i|\pi(e_i))}{\prod_{x_j\in\mathbf{x}}\Pr'_\mathbf{e}(x_j|\pi_\mathbf{e}(x_j))}$
$-\ln \Pr(\mathbf{e}) = \sum_{Cfg(\mathbf{X})} \Pr(\mathbf{x} \mid \mathbf{e}) \sum_{x_j\in\mathbf{x}} \ln \frac{\Pr(x_j|\pi(x_j))}{\Pr'_\mathbf{e}(x_j|\pi_\mathbf{e}(x_j))}$
$+ \sum_{Cfg(\mathbf{X})} \Pr(\mathbf{x}\mid\mathbf{e})\ln\prod_{e_i\in\mathbf{e}}\Pr(e_i \mid \pi(e_i)) - \ln\Pr(\mathbf{e})$
$= \sum_{X_j\in\mathbf{X}} \sum_{Cfg(X_j,\pi(X_j))} \ln \frac{\Pr(x_j|\pi(x_j))}{\Pr'_\mathbf{e}(x_j|\pi_\mathbf{e}(x_j))}$
$\sum_{Cfg(\mathbf{X}\setminus\{X_j,\pi(X_j)\})} \Pr(\mathbf{x} \mid \mathbf{e}) + \sum_{Cfg(\boldsymbol{\pi}(\mathbf{E}))}$
$\Pr(\boldsymbol{\pi}(\mathbf{e}) \mid \mathbf{e}) \ln \prod_{e_i\in\mathbf{e}} \Pr(e_i \mid \pi(e_i)) - \ln \Pr(\mathbf{e}) =$
$\sum_{X_j\in\mathbf{X}} \sum_{Cfg(X_j,\pi(X_j))} \Pr(x_j, \pi(x_j) \mid \mathbf{e}) \ln \frac{\Pr(x_j|\pi(x_j))}{\Pr'_\mathbf{e}(x_j|\pi_\mathbf{e}(x_j))}$
$+ \sum_{Cfg(\boldsymbol{\pi}(\mathbf{E}))} \Pr(\boldsymbol{\pi}(\mathbf{e}) \mid \mathbf{e}) \ln \prod_{e_i\in\mathbf{e}} \Pr(e_i \mid \pi(e_i))$
$-\ln\Pr(\mathbf{e}) = \sum_{X_j\in\mathbf{X}} \sum_{Cfg(X_j,\pi(X_j))} \Pr(x_j, \pi(x_j) \mid \mathbf{e})$
$\ln \Pr(x_j \mid \pi(x_j)) \qquad + \sum_{X_j\in\mathbf{X}} \sum_{Cfg(X_j,\pi(X_j))}$
$\Pr(x_j, \pi(x_j) \mid \mathbf{e}) \ln \frac{1}{\Pr'_\mathbf{e}(x_j|\pi_\mathbf{e}(x_j))}$
$+ \sum_{Cfg(\boldsymbol{\pi}(\mathbf{E}))} \Pr(\boldsymbol{\pi}(\mathbf{e}) \mid \mathbf{e}) \ln \prod_{e_i\in\mathbf{e}} \Pr(e_i \mid \pi(e_i))$
$-\ln \Pr(\mathbf{e})$ \hfill (Eq. 1)

The first, third, and fourth terms in Eq. 1 are decided by the original probability distribution, so $\sum_{X_j\in\mathbf{X}} \sum_{Cfg(X_j,\pi(X_j))} \Pr(x_j, \pi(x_j) \mid \mathbf{e}) \ln \Pr(x_j \mid \pi(x_j))$
$+ \sum_{Cfg(\boldsymbol{\pi}(\mathbf{E}))} \Pr(\boldsymbol{\pi}(\mathbf{e}) \mid \mathbf{e}) \ln \prod_{e_i\in\mathbf{e}} \Pr(e_i \mid \pi(e_i))$
$-\ln \Pr(\mathbf{e})$ is constant. To minimize the difference between *posterior* $BN_\mathbf{e}$ and $ESBN_\mathbf{e}$ the only choice is to minimize the following term:

$\sum_{X_j\in\mathbf{X}} \sum_{Cfg(X_j,\pi(X_j))} \Pr(x_j, \pi(x_j) \mid \mathbf{e}) \ln \frac{1}{\Pr'_\mathbf{e}(x_j|\pi_\mathbf{e}(x_j))}$
$= \sum_{X_j\in\mathbf{X}} \sum_{Cfg(\pi(X_j))} \sum_{Cfg(X_j)} \Pr(x_j, \pi(x_j) \mid \mathbf{e})$
$\ln \frac{1}{\Pr'_\mathbf{e}(x_j|\pi_\mathbf{e}(x_j))}$

This is equal to minimizing the term for each $X_j \in \mathbf{X}$ and each possible configuration of $\pi(x_j)$. $\sum_{Cfg(X_j)} \Pr(x_j, \pi(x_j) \mid \mathbf{e}) \ln \frac{1}{\Pr'_\mathbf{e}(x_j|\pi_\mathbf{e}(x_j))} =$
$\Pr(\pi(x_j) \mid \mathbf{e}) \sum_{Cfg(X_j)} \Pr(x_j \mid \pi(x_j), \mathbf{e}) \ln \frac{1}{\Pr'_\mathbf{e}(x_j|\pi(x_j)\setminus\mathbf{e})}$.
We have $\sum_{Cfg(X_j)} \Pr'_\mathbf{e}(x_j \mid \pi_\mathbf{e}(x_j))$
$= \sum_{Cfg(X_j)} \Pr'_\mathbf{e}(x_j \mid \pi(x_j) \setminus \mathbf{e}) = 1$. According to Shannon's information theory [Shannon, 1956], to minimize $\sum_{Cfg(X_j)} \Pr(x_j \mid \pi(x_j), \mathbf{e}) \ln \frac{1}{\Pr'_\mathbf{e}(x_j|\pi(x_j)\setminus\mathbf{e})}$ we should set $\Pr'_\mathbf{e}(x_j \mid \pi(x_j) \setminus \mathbf{e}) = \Pr(x_j \mid \pi(x_j), \mathbf{e})$. This proves the Theorem 1. □

## B  Proof of Corollary 1

**Proof.** Let $\mathbf{X} = \mathbf{V} \setminus \mathbf{E}$, then
$\sum_{Cfg(\mathbf{V}\setminus\mathbf{E})} \Pr_\mathbf{e}(\mathbf{v} \setminus \mathbf{e}) \ln \frac{\Pr_\mathbf{e}(\mathbf{v}\setminus\mathbf{e})}{\Pr'_\mathbf{e}(\mathbf{v}\setminus\mathbf{e})} =$
$\sum_{Cfg(\mathbf{X})} \Pr_\mathbf{e}(\mathbf{x} \mid \mathbf{e}) \ln \frac{\Pr(\mathbf{x}|\mathbf{e})}{\Pr'_\mathbf{e}(\mathbf{x})} =$
$\sum_{Cfg(\mathbf{X})} \Pr(\mathbf{x} \mid \mathbf{e}) \ln \frac{\prod_{x_j\in\mathbf{x}}\Pr(x_j|\pi(x_j))\prod_{e_i\in\mathbf{e}}\Pr(e_i|\pi(e_i))}{\Pr(\mathbf{e})\prod_{x_j\in\mathbf{x}}\Pr'_\mathbf{e}(x_j|\pi_\mathbf{e}(x_j))}$.
Since $\boldsymbol{\pi}(\mathbf{E}) \cap (\mathbf{V}\setminus\mathbf{E}) = \emptyset \Rightarrow \forall X_j \in \mathbf{X}, \Pr(x_j \mid \pi(x_j), \mathbf{e}) = \Pr(x_j \mid \pi(x_j))$, from Theorem 1, set $\Pr'_\mathbf{e}(x_j \mid \pi_\mathbf{e}(x_j)) = \Pr(x_j \mid \pi(x_j))$ to minimize the divergence, then
$\sum_{Cfg(\mathbf{X})} \Pr(\mathbf{x} \mid \mathbf{e}) \ln \frac{\prod_{x_j\in\mathbf{x}}\Pr(x_j|\pi(x_j))\prod_{e_i\in\mathbf{e}}\Pr(e_i|\pi(e_i))}{\Pr(\mathbf{e})\prod_{x_j\in\mathbf{x}}\Pr'_\mathbf{e}(x_j|\pi_\mathbf{e}(x_j))} =$
$\sum_{Cfg(\mathbf{X})} \Pr(\mathbf{x} \mid \mathbf{e}) \ln \frac{\prod_{e_i\in\mathbf{e}}\Pr(e_i|\pi(e_i))}{\Pr(\mathbf{e})}$. Also from $\boldsymbol{\pi}(\mathbf{E}) \cap (\mathbf{V}\setminus\mathbf{E}) = \emptyset$, $\forall E_i \in \mathbf{E}$, $\pi(E_i) \subseteq \mathbf{E} \Rightarrow \prod_{e_i\in\mathbf{e}}\Pr(e_i \mid \pi(e_i)) = \Pr(\mathbf{e})$, so $\sum_{Cfg(\mathbf{X})} \Pr(\mathbf{x} \mid \mathbf{e}) \ln \frac{\prod_{e_i\in\mathbf{e}}\Pr(e_i|\pi(e_i))}{\Pr(\mathbf{e})} = 0$. The KL-divergence between $\Pr_\mathbf{e}$ and $\Pr'_\mathbf{e}$ is zero, thus $\Pr_\mathbf{e}(\mathbf{v} \setminus \mathbf{e}) = \Pr'_\mathbf{e}(\mathbf{v} \setminus \mathbf{e})$ according to [Kullback, 1959]. □

## C  Proof of Lemma 2

**Proof.** $\forall X_k \in Ah(X_j) \setminus \beta_e(X_j)$, consider the following three cases.

Case 1: If a path $X_k \to E$ exists then we show that this path is d-separated by $\alpha_e(X_j)$. There are two possibilities. First, $X_k \to E$ bypasses $X_j$, so it must pass one of the parents of $X_j$. Then $\pi(X_j)$ d-separates the path. Second, $X_k \to E$ does not pass $X_j$. Then the path must be d-separated by $\alpha_e(X_j) \setminus X_j$, so $(\alpha_e(X_j) \setminus X_j) \cup \pi(X_j)$ d-separates the path.

Case 2: If paths $N \to X_k$ and $N \to E$ exist, so $N \in Ah(X_j)$, and $N$ d-separate the $X_k$ and $E$, according to Case 1, $(\alpha_e(X_j)\setminus X_j)\cup\pi(X_j)$ d-separates path $N \to E$.

Case 3: If paths $X_k \to B$ and $E \to B$ exist, according to topological order $\{B, descendants\,of\,B\} \cap ((\alpha_e(X_j)\setminus X_j)\cup\pi(X_j)) = \emptyset$, so $(\alpha_e(X_j)\setminus X_j)\cup\pi(X_j)$ d-separates this path.

From cases 1 to 3 we see that $\langle E, Ah(X_j) \mid ((\alpha_e(X_j)\setminus X_j)\cup\pi(X_j))\rangle$, so $\beta_e(X_j) \subseteq (\alpha_e(X_j) \setminus X_j) \cup \pi(X_j)$. □